%% file: mypaper.tex
\documentclass[conference]{IEEEtran}
\IEEEoverridecommandlockouts
\usepackage{cite}
\usepackage{amsmath,amssymb,amsfonts}
\usepackage{algorithmic}
\usepackage{graphicx}
\usepackage{textcomp}
\usepackage{xcolor}
\usepackage{underscore}
\usepackage{booktabs}
\usepackage{multirow}
\usepackage{url}
\usepackage{threeparttable}
\def\BibTeX{{\rm B\kern-.05em{\sc i\kern-.025em b}\kern-.08em
		T\kern-.1667em\lower.7ex\hbox{E}\kern-.125emX}}
	
\begin{document}
	
	\title{Entity-Aware Self-Attention and Contextualized GCN for Enhanced Relation Extraction in Long Sentences
	}
	\author{
		\IEEEauthorblockN{
			Xin Wang,
			Xinyi Bai 
			}
		\IEEEauthorblockA{
			Henan Institute of Technology, China \\
	}}
	\maketitle
	
	\begin{abstract}
		Relation extraction as an important natural Language processing (NLP) task is to identify relations between named entities in text. Recently, graph convolutional networks over dependency trees have been widely used to capture syntactic features and achieved attractive performance. However, most existing dependency-based approaches ignore the positive influence of the words outside the dependency trees, sometimes conveying rich and useful information on relation extraction. In this paper, we propose a novel model, Entity-aware Self-attention Contextualized GCN (ESC-GCN), which efficiently incorporates syntactic structure of input sentences and semantic context of sequences. To be specific, relative position self-attention obtains the overall semantic pairwise correlation related to word position, and contextualized graph convolutional networks capture rich intra-sentence dependencies between words by adequately pruning operations. Furthermore, entity-aware attention layer dynamically selects which token is more decisive to make final relation prediction. In this way, our proposed model not only reduces the noisy impact from dependency trees, but also obtains easily-ignored entity-related semantic representation. Extensive experiments on various tasks demonstrate that our model achieves encouraging performance as compared to existing dependency-based and sequence-based models. Specially, our model excels in extracting relations between entities of long sentences.
	\end{abstract}
	
	\begin{IEEEkeywords}
		relation extraction, self-attention, dependency trees, semantic representation
	\end{IEEEkeywords}

	\section{Introduction}
		There has been major interest in relation extraction, which aims to assign a relation among a pair of entity mentions from plain text. Relation extraction is the basis for answering knowledge queries \cite{yu2017improved,yin2022generic}, building knowledge base \cite{sorokin2017context,yin2022dynamic}, and also forming an important supporting technology for information extraction \cite{qian2019graph,yin2022deal}. Recent models for relation extraction are primarily built on deep neural networks, which encode the entire sentence to obtain relation representations and have made great progress \cite{zeng2014relation, zhang2015relation}.
		\par
\begin{figure*}
			\centering
			\includegraphics[width=0.86\textwidth]{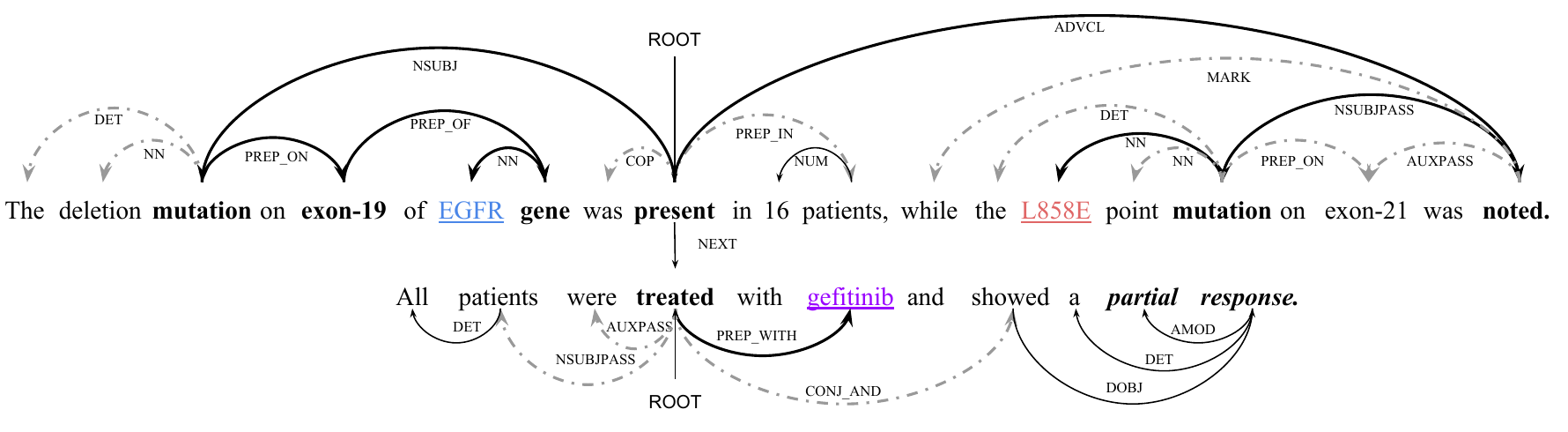}
			\caption{Example of dependency parsing for two sentences expressing a ternary interaction. The SDP between three entities (identify with different color) is highlighted in bold (edges and tokens). The root node of the LCA subtree of entities is \textit{present}. The dotted edges indicate tokens $K$=1 away from the subtree. Tokens ``\textit{partial response}'' are off these two paths.}
			\label{figure:deptree}
		\end{figure*}

 From the example given in Fig.~\ref{figure:deptree} for cross-sentence $n$-ary task, there is a relation ``\textbf{\textit{sensitivity}}'' between the three entities within the two sentences, which expresses that ``tumors with \textit{L858E} mutation in \textit{EGFR} gene respond to \textit{gefitinib} treatment''. The edges connecting different tokens identify their dependency labels. Prior efforts show that models utilizing dependency parsing of input sentences (i.e., dependency-based models) are very effective in relation extraction, because their superiority lies in drawing direct connections between distant syntactically correlated words. Xu et al.\cite{xu2015classifying} first applied LSTM on the shortest dependency path (SDP) between the entities in the full tree. Miwa et al.\cite{miwa2016end} reduced the full tree into the subtree below the lowest common ancestor (LCA) of the entities. Both patterns prune the dependency trees between the entities to cover relevant information and discard noises. However, if only the dependency structure (i.e., SDP, LCA) shown in Fig.~\ref{figure:deptree} is considered, the tokens ``\textit{\textbf{partial response}}'' will be neglected, yet, they contribute to the gold relation greatly. Therefore, it is very essential to obtain the interactions of all words, not just the dependency trees of entities. To address this issue, we use a relative position self-attention mechanism, which allows each token to take its left and right context into account while calculating pairwise interaction scores with other tokens.
		
		\par Recently, combining entity position features with neural networks has greatly improved the performance of relation extraction. Zhang et al.\cite{zhang2017position} combined sequence LSTM model with a position-attention mechanism and got a competitive results. Lee et al.\cite{lee2019semantic} proposed a novel entity-aware BiLSTM with Latent Entity Typing (LET), and obtained state-of-the-art performance on SemEval dataset. From their experiments, we conclude that the words that determine the relation frequently related to the target entities. However, these methods only utilize the semantic representations and position features, ignoring the dependency syntax of the words. Unlike previous efforts, which focus on either dependency parsing or the semantic features, we synthesize syntactic dependency structure and entity-related sequential semantic context into an attention mechanism, both of which are crucial for relation extraction.
		\par In this paper, we first utilize relative position self-attention mechanism to encode  semantic interactions of the sequential sentence, which ignores the distance between words to calculate the compatibility scores and relative position scores. Then contextualized graph convolution module encodes the dependency trees between the entities to capture contextual long-range dependencies of words. Afterwards, entity-aware attention mechanism combines these two modules to get final relation representations. The contributions of our work can be summarized as follows: 
		\\1) We propose a ESC-GCN model to learn relation representations. Compared with previous methods, our method not only utilizes semantic features but also considers dependency features. 
		\\2) Our proposed model prove to be very competitive on the sentence-level task (i.e., TACRED and SemEval dataset) and cross-sentence $n$-ary task. Especially, our model outperforms most baseline in long sentences. 
		\\3) We show that our model is interpretable by visualizing the relative position self-attention.

	\section{Related Work}
		Recently, deep neural models have shown superior performances in the field of NLP. Compared with traditional hand-crafted models, deep neural models can automatically learn latent features and greatly improve performances~\cite{zhang2022would,zhang2023investigating,meng2024deep}.
		\par Relation extraction has been intensively studied in a long history, and most existing neural relation extraction models can be divided into two categories: sequence-based and dependency-based. Zeng et al.\cite{zeng2014relation} first applied CNN with manual features to encode relations. Wang et al.\cite{wang2016relation} proposed attention-based CNN, which utilizes word-level attention to better determine which parts of the sentence are more influential. Variants of Convolutional Neural Networks (CNNs) methods have been proposed, including CNN-PE \cite{nguyen2015relation,yin2024continuous} , CR-CNN \cite{santos2015classifying} and Attention-CNN\cite{shen2016attention}. Besides CNN-based architecture, the RNN-based models are another effective approaches. Zhang et al.\cite{zhang2015relation} first applied RNN to relation extraction and got competitive performance. Zhang et al.\cite{zhang2015bidirectional} employed BiLSTM to learn long-term dependencies between entity pairs. Moreover, variants RNNs methods such as Attention-LSTM \cite{zhou2016attention,yin2024dynamic} and Entity-aware LSTM \cite{lee2019semantic} have been proposed. However, these models only considered the sequential representations of sentences and ignored the syntactic structure. Actually, these two features complement each other. By combining C-GCN \cite{zhang2018graph,yinsport} and PA-LSTM \cite{zhang2017position}, Zhang et al.\cite{zhang2018graph} obtained better results on TACRED dataset.
		\par Compared with the sequence-based models, incorporating dependency syntax into neural models has proven to be more successful, which captures non-local syntactic relations that are only implicit in the surface from alone\cite{ai2023gcn,shou2023adversarial,tang2024merging, xu2016improved,ju2024survey}. Xu et al.\cite{xu2015classifying} proposed SDP-LSTM that leverages the shortest dependency path between two entities. Miwa et al.\cite{miwa2016end} reduced the full tree into the subtree below the lowest common ancestor, which combined a Tree-LSTM \cite{tai2015improved} and BiLSTMs on tree structures to model jointly entity and relation extraction. Peng et al.\cite{peng2017cross} proposed a graph-structured LSTM for cross-sentence $n$-ary relation extraction, which applied two directed acyclic graphs (DAGs) LSTM to capture inter-dependencies in multiple sentences. Song et al.\cite{song2018n} proposed a graph-state LSTM model which employed a parallel state to model each word, enriching state scores via message passing. Zhang et al.\cite{zhang2018graph} presented C-GCN for relation extraction, which uses graph convolution and a path-centric pruning strategy to selectively include relative information. Vashishth et al.\cite{vashishth2018incorporating,yin2023dream} utilized GCN to incorporate syntactic and semantic information of sentences to learn its word embedding. Sahu et al.\cite{sahu2020relation} proposed Self-determined GCN (S-GCN) which determines a weighted graph using a self-attention mechanism.
		\par Vaswani et al.\cite{vaswani2017attention} proposed an attention-based model called Transformer. It's a mainstream that combined attention mechanism with CNNs \cite{zeng2014relation,yin2023omg,yin2023messages} or RNNs \cite{zhang2015relation,pang2023sa} in the past few years. Recently, attention mechanism has been proven to capture helpful information for relation extraction \cite{shen2016attention,yin2023coco}. Tran et al.\cite{tran2019relation} utilized Segment-Level Attention-based CNN and Dependency-based RNN for relation classification, which got a comparable result on SemEval dataset. Verga et al.\cite{verga2018simultaneously} used self-attention to encode long contexts spanning multiple sentences for biological relation extraction. Zhang et al.\cite{zhang2017position} employed a position-attention mechanism over LSTM outputs for improving relation extraction. Bilan et al.\cite{bilan2018position} substituted the LSTM layer with the self-attention encoder for relation extraction. Yu et al.\cite{yu2019beyond} proposed a novel segment attention layer for relation extraction, and achieved competitive results on TACRED dataset.
		
	\begin{figure*}
		\centering
		\includegraphics[width=0.83\textwidth]{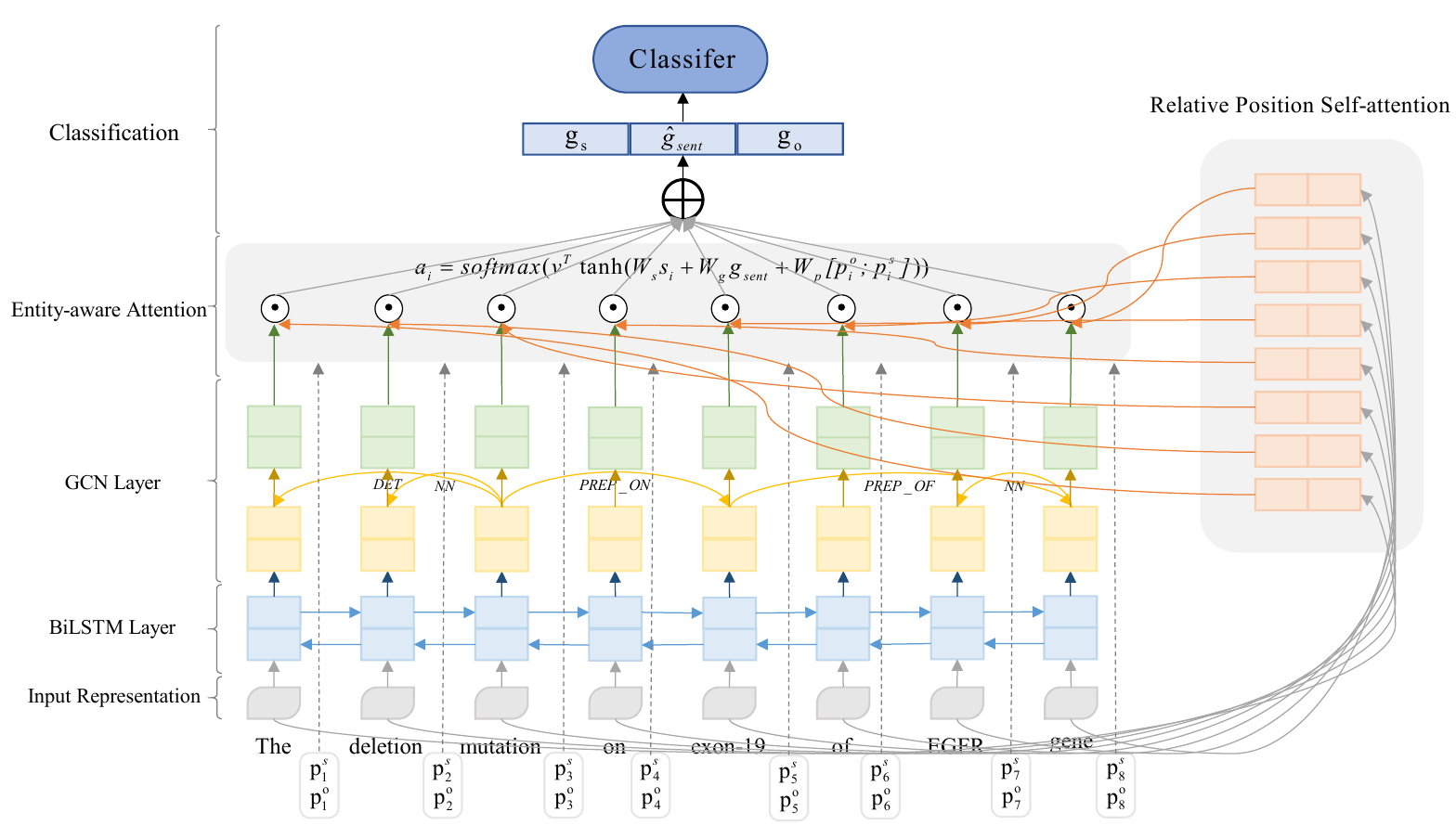}
		\caption{Overview of our proposed ESC-GCN model, an example sentence ``The deletion mutation on exon-19 of EFGR gene''.}
		\label{figure:esc-gcn}
	\end{figure*}
	
	\section{The Proposed Model}
		Following the existing studies \cite{zhang2018graph}, we define relation extraction as a multi-class classification problem, which can be formalized as follows: Let $S = \left[w_{1}, w_{2}, \ldots, w_{n}\right]$ denote a sentence, where $w_{i}$ is the $i$-th token. A subject entity and an object entity are identified: $W_{s} = \left[w_{s_{1}}, w_{s_{2}}, \ldots, w_{s_{n}}\right]$ and $W_{o} = \left[w_{o_{1}}, w_{o_{2}}, \ldots, w_{o_{n}}\right]$. Given $S$, $W_{s}$ and $W_{o}$, the goal of relation extraction is to predict a relation $r \in R$ or “\textit{no relation}”.
		\par Specifically, Fig.~\ref{figure:esc-gcn} depicts the overall architecture of our model, which contains the following four modules: (i) input representation, which encodes original sentence into a sequence of vectors and get a positional vector of each token; (ii) relative position self-attention mechanism, which combines relative position embedding and self-attention to obtain more powerful representations; (iii) contextualized GCN layer that performs graph convolution over pruning dependency trees following by BiLSTM; (iiii) classification module obtains hidden representation of all tokens with entity-aware attention to predict a relation among entities.

	\subsection{Preliminary}
		GCNs are neural networks that operate directly on graph structures \cite{kipf2016semi}, which are an adaptation of convolutional networks. Given a graph with $n$ nodes, we generate the graph with an $n \times n$ adjacency matrix A where $A_{ij}$ = 1 if there is an edge going from node $i$ to node $j$, otherwise $A_{ij}$ = 0. Similar to Marcheggiani et al.\cite{marcheggiani2018exploiting}, we extend GCNs for encoding dependency trees by incorporating opposite of edges into the model. Each GCN layer takes the node embedding from the previous layer $g_{j}^{(l-1)}$ and the adjacency matrix $A_{ij}$ as input, and outputs updated node representation for node $i$ at the $l$-th layer. Mathematically, the induced representation $g_{i}^{(l)}$ can be defined as :
	\begin{equation}
		g_{i}^{(l)}=\rho\left(\sum_{j=1}^{n}A_{ij}W^{(l)}g_{j}^{(l-1)}+b^{(l)}\right), \label{equation1}
	\end{equation}
		where $W^{(l)}$ is a linear transformation, $b^{(l)}$ is the bias vector, and $\rho$ is an activation function (e.g., RELU). $g_{i}^{(0)}$ is the initial input $h_{i}^{(L_{1})}$, more details can be seen from subsection \ref{subsection:cgcn}.
		
	\subsection{Input Representation}\label{subsection:input}
		Distributed representation of words in a vector space is helpful to achieve better performance in NLP tasks. Accordingly, we embed both semantic information and positional information of words into their input embeddings, respectively.
		\par In our model, the input representation module first transforms each input token $w_{i}$ into a comprehensive embedding vector $x_{i}$ by concatenating its word embedding $word_{i}$, entity type embedding $ner_{i}$ and part-of-speech (POS) tagging embedding $pos_{i}$. Embedding vector $x_{i}$ formally defined as follow:
		\begin{equation}
			x_{i} = [word_{i};ner_{i};pos_{i}].
		\end{equation}
		\par It has been proved that the words close to the target entities are usually more informative in determining the relation between entities \cite{bilan2018position}, we modify the position representation originally proposed by \cite{zhang2017position}, and convert it into binary position encoding. Consequently, we define a binary-position sequence $\left[p_{1}^{s}, \ldots, p_{n}^{s}\right]$ that relative to the subject entity:
		\begin{equation}
			p_{i}^{s}=\left\{\begin{array}{lcl}
				-\lfloor\log_{2}(s_{1}-i)\rfloor-1, & i<s_{1} \\
				0, & s_{1} \leq i \leq s_{2} \\
				\lfloor\log_{2}(i-s_{2})\rfloor+1, & i>s_{2}
			\end{array}\right.,
		\end{equation}
		where $s_{1}$, $s_{2}$ represent the start index and end index of the subject entity respectively, $p_{i}^{s} \in \mathbb{Z}$ can be viewed as the relative distance of token $x_{i}$ to the
		subject entity. 
		\par Similarly, we also obtain a position sequence $\left[p_{1}^{o}, \ldots, p_{n}^{o}\right]$ relative to the object entities. By concatenating the position embeddings, we get a unified position embedding $p_{i} \in \mathbb{R}^{d_{p} \times 2}$, and $d_{p}$ indicates the dimension of position embedding.

		\begin{figure}
			\centering
			\includegraphics[scale=0.55]{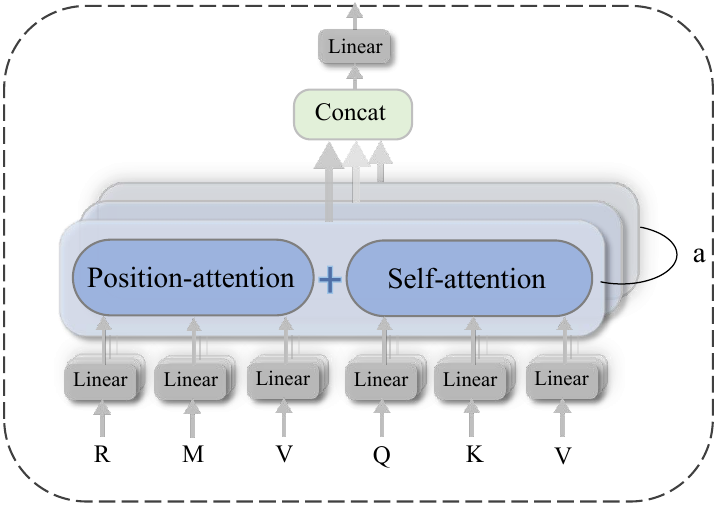}
			\caption{Relative Position self-attention structure.}
			\label{figure:Pattn-Sattn}
		\end{figure}
	
	\subsection{Relative Position Self-attention Mechanism}
		The self-attention mechanism proposed by Vaswani et al.\cite{vaswani2017attention}, which allows words to take its context into account. Following Bilan et al.\cite{zhang2017position}, we apply several modifications to the original self-attention layer. Firstly, we simplify the residual connection that directly goes from the self-attention block to the normalization layer. Then we substitute the layer normalization with batch normalization. In our experiments, we have observed improvements with these setting, and a more detailed overview of the results can be seen in the subsection \ref{subsection:ablation}. 
		\par Traditionally, a self-attention layer takes a word representation at position $i$ as the query (a matrix $Q$ holds the queries for position $i$) and computes a compatibility score with representations at all other positions (represented by a matrix $V$). The score w.r.t. position $i$ is reformed to an attention distribution over the entire sentence, which is used as a weighted average of representations $E$ at all positions. 
		\par Shaw et al.\cite{shaw2018self} exploited the relative positional encoding to improve the performance of self-attention. Similarly, we modify our self-attention layer, together with a position attention that takes into account positions of the query and the object in the sentence. For one attention head $a$, our self-attention head $s_{i}^{(a)}$ obtain its representation by summing pairwise interaction scores and relative position scores together, formally defined as follows:
	\begin{equation}
		s_{i}^{(a)} = softmax(\frac{QK^{T}+RM^{T}}{\sqrt{d_{w}}})V,
	\end{equation}
		where $Q=W^{q(a)} e_{i}$, $K=W^{k(a)} E$, $V=W^{v(a)} E$ wherein $W^{q(a)}, W^{k(a)}, W^{v(a)}$ are linear transformations, which map the input representation into lower-dimensional space. $M$ is relative position embedding matrix:
	\begin{equation}
		M_{i}=\left[m_{1-i}, \ldots, m_{-1}, m_{0}, m_{1}, \ldots, m_{n-i}\right],
	\end{equation}
		where $n$ is the length of the input sentence and the matrix $M_{i}$ is the relative position vectors, $m_{0}$ is at position $i$ and other $m_{j}$ are ordered relative to position $i$. 
		\par Similar to $Q$, we obtain a query vector $R=W^{r(a)} e_{i}$ to obtain position relevance. The position attention scores result from the interaction of $R$ with the relative position vectors in $M_{i}$. As show in Fig.~\ref{figure:Pattn-Sattn}, we associate position attention scores with the pairwise interaction scores, which incorporates position features into overall dependencies of sequence.

	\subsection{Contextualized GCN Layer} \label{subsection:cgcn}
		In this section, we construct a contextualized GCN model which takes the output from subsection \ref{subsection:input} as input $h^{(0)}$ of this module. A BiLSTM layer is adopted to acquire the context of sentence for each word $w_{i}$. For explicitly, we denote the operation of LSTM unit as LSTM($x_{i}$). The contextualized word representations is obtained as follows:
	\begin{equation}
		h_{i}=\left[\overrightarrow{\operatorname{LSTM}}\left(x_{i}\right) ; \overleftarrow{\operatorname{LSTM}}\left(x_{i}\right)\right], i \in[1, n],
	\end{equation}
		where $h_{i} \in \mathbb{R}^{2 \times d_{h}}$ and $d_{h}$ indicates the dimension of LSTM hidden state. Then we obtain hidden representations of all tokens $h^{(L_{1})}$, which represents the input $g^{(0)}$ for graph convolution, where $L_{1}$ represents the layer number of RNN.
		\par The GCN model \cite{kipf2016semi} has been popularly used for learning graph representation. Dependency syntax has been recognized as a crucial source of features for relation extraction \cite{liu2015dependency}, and most of the information involved relation within the subtree rooted at the LCA of the entities. Miwa et al.\cite{miwa2016end} has shown that discarding those noises outside of LCA can help relation extraction. Before applying graph convolution operation, we do some tricks on the dependency parsing tree, which keeps the original dependency path in the LCA tree and incorporates 1-hop dependencies away from the subtree. Accordingly, we cover the most relevant content and remove irrelevant noise as much as possible.
		\par Originally applying the graph convolution in (\ref{equation1}) could bring about node representations with obviously different scale \cite{zhang2018graph}, since the degree of tokens varies a lot. Furthermore, Equation (\ref{equation1}) is never carried the nodes themselves. To cope with the above limitations, we resolve these issues by normalizing the activations in the graph convolution, and add self-loop into each node in adjacency matrix A, modified graph convolution operation as follows:
	\begin{equation}
		g_{i}^{(l)}=\sigma\left(\sum_{j=1}^{n} \tilde{A}_{i j} W^{(l)} g_{j}^{(l-1)} / d_{i}+b^{(l)}\right),
	\end{equation}
		where $\tilde{A}=A+I$, $I$ is the $n \times n$ identity matrix, and $d_{i}=\sum_{j=1}^{n} \tilde{A}_{i j}$ is the degree of token $i$. This operation updates the representation of node $i$ by aggregating its neighborhood via a convolution kernel. After $L_{2}$ iterations, we obtain the hidden outputs of graph convolution $g^{(L_{2})}$, where $L_{2}$ represents the layer number of GCN.

	\subsection{ESC-GCN for Relation Extraction}
		After applying the $L_{2}$-layer contextualized GCN model, we obtain hidden representation of each token, which is directly influenced by its neighbors (no more than $L_{2}$ edges apart in the dependency trees). To make use of graph convolution for relation extraction, we first obtain a sentence representation as follows:
	\begin{equation}
		g_{sent}=f\left(g^{(L_{2})}\right)=f\left(\operatorname{GCN}\left(g^{(0)}\right)\right),
	\end{equation}
		where $g^{(L_{2})}$ denotes the collective hidden representation at layer $L_{2}$ of the GCN, and $f: \mathbb{R}^{d \times n} \rightarrow \mathbb{R}^{d}$ is a max pooling function that maps from $n$ output vectors to the sentence vector. Moreover, we also obtain a subject representation $g_{s}$ as follows:
	\begin{equation}
		g_{s}=f\left(g_{s_{1}:s_{2}}^{(L_{2})}\right),
	\end{equation}
		as well as an object representation $g_{o}$ respectively. 
		\par The final computation of the entity-aware attention utilizes the output state of GCN (i.e., a summary vector $g_{sent}$), the self-attention hidden states output vector $s_{i}$, and the embeddings for the subject and object relative positional vectors $p_{i}^{s}$, $p_{i}^{o}$. For each hidden state $s_{i}$, an attention weight $\alpha_{i}$ is calculated using the following two equations:
	\begin{equation}
		u_{i}=v^{\top} \tanh \left(W_{s} s_{i}+W_{g} g_{sent}+W_{p}[p_{i}^{s};p_{i}^{o}] \right),
	\end{equation}
	\begin{equation}
		\alpha_{i}=\frac{\exp \left(v^{\top} u_{i}\right)}{\sum_{j=1}^{n} \exp \left(v^{\top} u_{j}\right)},
	\end{equation}
		where $W_{s}$ weights are learned parameters using self-attention, $W_{g}$ weights are learned parameters using contextualized GCN and $W_{p}$ weights are learned using the positional encoding embeddings. 
		\par Afterwards, $\alpha_{i}$ is used to convert the information that combines relative position self-attention and long-distance dependency relationship, which decides on how much each GCN outputs should contribute to the final sentence representation $\hat{g}_{sent}$ as follows:
	\begin{equation}
		\hat{g}_{sent} = \sum_{i=1}^{n} \alpha_{i} g_{i}^{(L_{2})}.
	\end{equation}
		\par Then the representation $\hat{g}_{sent}$, $g_{s}$ and $g_{o}$ are concatenated and fed into a feed-forward neural network (FFNN):
	\begin{equation}
		g_{final}=\text{FFNN}\left(\left[\hat{g}_{sent} ; g_{s} ; g_{o}\right]\right).
	\end{equation}
		\par In the end, the final sentence representation $g_{final}$ is then fed to another MLP layer followed by a softmax operation to obtain a probability distribution over relations:
	\begin{equation}
		p(r \mid g_{final})=\operatorname{softmax}(w \cdot g_{final}+b),
	\end{equation}
		where $g_{final}$ is the sentence representation, and $r$ is the target relation, $w$ is a linear transformation and $b$ is a bias term. We utilize the cross entropy and the L2 regularization to define the objective function as follows:
	\begin{equation}
		J(\theta)=-\sum_{i=1}^{s}\left(y_{i} \mid x_{i}, \theta\right)+\beta\|\theta\|^{2},
	\end{equation}
		where $s$ indicates the total sentence; $x_{i}$ and $y_{i}$ represent the sentence and relation label of the $i^{th}$ training example; $\beta$ is $L_{2}$ regularization hyper-parameter. The $\theta$ is the whole network parameter, which can be learnable.

	\input{table/crossdataset.tex}
	\input{table/sentencedataset.tex}
	
 	\section{Experiments}
 		In this section, we evaluate our ESC-GCN model with four datasets on two tasks, namely cross-sentence $n$-ary relation extraction and sentence-level relation extraction.
	
	\input{table/cross.tex}

 	\subsection{Dataset}
 		For cross-sentence $n$-ary relation extraction, we use two datasets generated by Peng et al.\cite{peng2017cross}, which is a biomedical-domain dataset focusing on drug-gene-mutation relations.\footnote{The number of entities is fixed in $n$-ary relation extraction task. It is 3 for the ternary and 2 for the binary.} It contains 6,987 ternary relation instances and 6,087 binary relation instances extracted from PubMed.\footnote{The dataset is available at {https://github.com/freesunshine0316/nary-grn}} Most instances contain multiple sentences and are assigned to one of the five categories, e.g. “resistance or non-response”, “sensitivity”, “response”, “resistance” and “None”. we define two sub-tasks, i.e., binary-class $n$-ary relation extraction and multi-class $n$-ary relation extraction. Table \ref{table:crossdataset} shows statistics of the dataset. 

 		\par For sentence-level relation extraction, we follow the experimental settings in Zhang et al.\cite{zhang2018graph} to evaluate our ESC-GCN model on the TACRED dataset \cite{zhang2017position} and Semeval-2010 Task 8 dataset \cite{hendrickx2019semeval}. TACRED contains over 106k mention pairs collected from the TACKBP evaluations 2009–2014. It includes 41 relation types and a ``$no\_relation$'' class when no relation is hold between entities. Mentions in TACRED are typed, subjects are classified into person and organization, and objects are categorized into 16 fine-grained classes(e.g., date, location, title). The SemEval-2010 Task 8 dataset is an acknowledged benchmark for relation extraction (1/10 of TACRED). The dataset defines 9 types of relations (all relations are directional) and a class ``$other$'' denoted no relation. There are 10,717 annotated sentences which consist of 8,000 samples for training and 2,717 samples for testing. Table \ref{table:sentencedataset} shows statistics of the dataset.
		
 	\subsection{Results on Cross-Sentence $n$-ary Relation Extraction}
		To ensure a fair comparison, the model is evaluated using the same metrics as Song et al.\cite{song2018n} for cross-sentence $n$-ary tasks, we consider three kinds of baselines: 1) a feature-based classifier \cite{quirk2016distant} which utilizes SDP between two entity pairs. Additionally, a tree-structured LSTM methods (SPTree) \cite{miwa2016end}, 2) Models extend LSTMs by encoding the graph structure of the dependency tree (i.e., Graph LSTM \cite{peng2017cross}, Bidirectional Directed Acyclic Graph LSTM (bidir DAG LSTM) \cite{song2018n}, Graph State LSTM (GS LSTM) \cite{song2018n}), 3) Graph convolutional networks (GCN) \cite{zhang2018graph} which has already proved effectively on the relation extraction. The five-fold cross validation results are shown in Table \ref{table:cross}, the column of the GCN shows the best results under different pruning strategies. For Binary-class ternary task (first two columns in Table \ref{table:cross}), our ESC-GCN achieves an accuracy of 86.2 and 86.7 under \texttt{Single} and \texttt{Cross} setting,  respectively, outperforming all baselines. For Binary-class binary task  (third and fourth columns in Table \ref{table:cross}), the ESC-GCN achieves accuracy of 85.0 and 84.6 under \texttt{Single} and \texttt{Cross} setting. Graph LSTM variants models tend to achieve higher results than feature-based models. Our ESC-GCN surpasses the overall of Graph LSTM variants, which has demonstrated graph convolution is more effective than the Graph LSTM. Intuitively, longer sentences in the multiple sequence correspond to more complex dependency structures. We notice that our ESC-GCN achieves a better test accuracy than C-GCN on Multi-class (2.5 and 1.2 points improvement), which further demonstrates its ability to learn better representation with entity features in multiple sentence setting.
		
		\input{table/tacred.tex}

 	\subsection{Results on Sentence-level Relation Extraction}
 		 For sentence-level task, we report the micro-averaged $F_{1}$ scores for the TACRED dataset and the macro-averaged $F_{1}$ scores for the SemEval-2010 task 8 dataset. we use the average test $F_{1}$ scores derived from five independently run models. We now report the results on the TACRED dataset in Table \ref{table:tacred}. we compare our model against following baselines: 1) sequence-based models, i.e., Convolutional Neural Networks (CNN-PE) \cite{nguyen2015relation}, Position Aware LSTM (PA-LSTM) \cite{zhang2017position}, Self-Attention Encoder (Self-Attn) \cite{bilan2018position}, Segment Attention LSTM (SA-LSTM) \cite{yu2019beyond}; 2) dependency-based models, i.e., the short dependency path LSTM (SDP-LSTM) \cite{xu2015classifying}, Tree-structured LSTM (Tree-LSTM) \cite{tai2015improved}; and 3) graph-based models: GCN and Contextualized GCN (C-GCN) \cite{zhang2018graph}, Simplifying Graph Convolutional Networks (S-GCN) \cite{wu2019simplifying}. As shown in Table \ref{table:tacred}, our ESC-GCN shows better performance than all baselines on the TACRED dataset, which achieves $F_{1}$ of 67.1, outperforming C-GCN by 0.7 $F_{1}$ points. This result shows the effectiveness of semantic representation. Our model obtains the highest precision, but get a general recall value. The performance gap between Self-Attn and ESC-GCN shows that our model is better at incorporating dependency relations and the context of entities in the sentence-level task.
 		 
 		 \input{table/semeval.tex}

 		 \par We also evaluate our model on the SemEval dataset. Table \ref{table:semeval} shows the experimental results. Apart from above baselines, we compare ESC-GCN with other sequence-based models, i.e., Attention-based BiLSTM (BiLSTM+Attn) \cite{zhou2016attention}, Entity-aware BiLSTM with a LET \cite{lee2019semantic} and dependency-based models, i.e., tree-structured LSTM methods (SPTree) \cite{miwa2016end}. Our ESC-GCN achieves a competitive performance ($F_{1}$-score of 85.1) on the SemEval dataset. The result shows that integrating entity features and dependency parsing can obtain better representation for relation extraction.
		
		\input{table/ablation.tex}
		
 	\section{Analysis} \label{section:analysis}
 	
 	\subsection{Ablation Study} \label{subsection:ablation}
 		In order to study the contribution of each component, we conducted an ablation study on the TACRED. Table \ref{table:ablation} shows the results. Instead of default residual connections described by \cite{vaswani2017attention}, the optimized residual connection contributes 0.2 points. Similarly, layer normalization contributes 0.4 points improvement. The entity-aware module and self-attention module contribute 0.5 and 0.7 points respectively, which illustrates that both layers promote our model to learn better relation representations. When we remove the feedforward layers and the entity representation, $F_{1}$ score drops by 0.9 points, showing the necessity of adopting ``multi-channel'' strategy. We also notice that the BiLSTM layer is very effective for relation extraction, which drops the performance mostly ($F_{1}$ relatively drops 2.3 points).

 	\subsection{Performance with different Pruned Trees}
		Table \ref{table:prune} shows the performance of the ESC-GCN model with different pruned trees on the TACRED dev set, where $K$ means that the pruned trees include tokens that are up to distance $K$ away from the dependency path in the LCA subtree. Specifically, we observe that our ESC-GCN achieves the highest $F_{1}$ of 67.1 with the setting as $K$ = 1, showing that incorporating 1-hop dependencies in the LCA greatly optimize the performance without introducing too much noise. In addition, we notice that the performance of our model drops a lot (relatively 1.1 points) with full trees, indicating too much extra dependency information will hurt the performance. When $K$ = 0, we only obtain the dependency trees between the entities that drops the performance of relation extraction.
		\subsection{Interpretation of self-attention}
\begin{figure}
 		\centering
 		\includegraphics[scale=0.35]{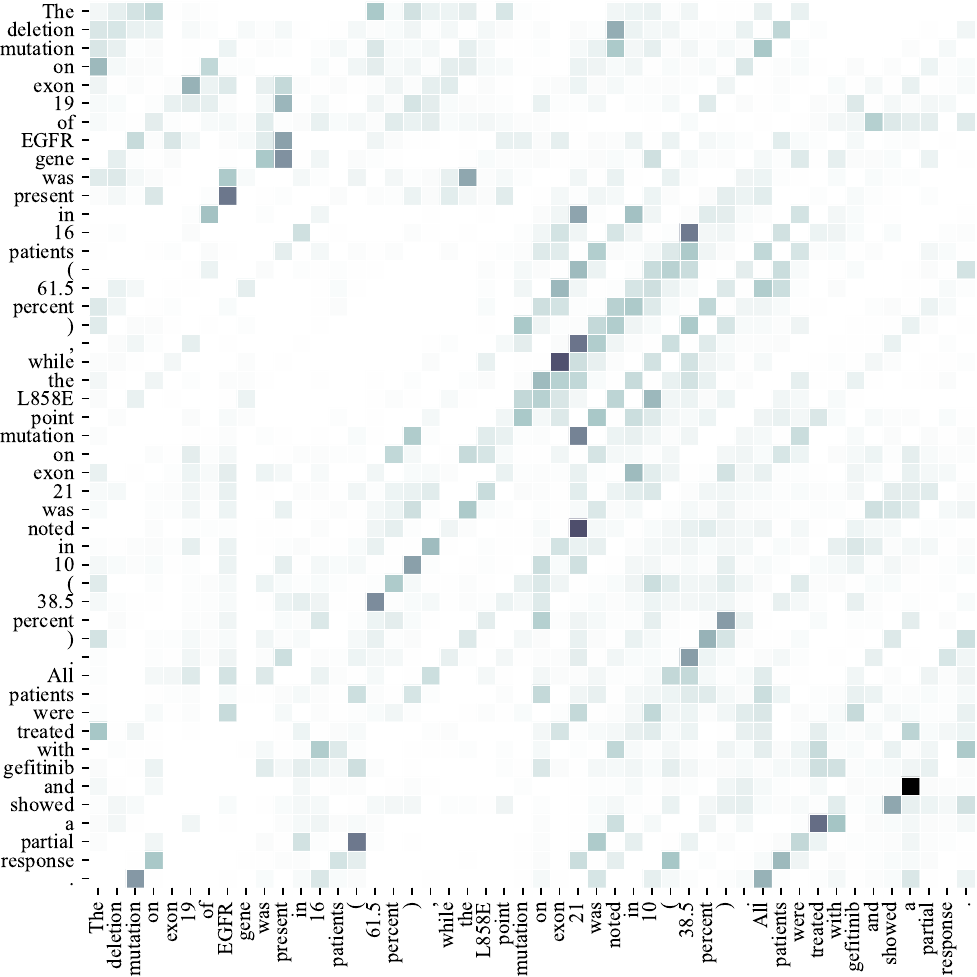}
 		\caption{Visualization of attention scores in the relative position self-attention layer. Darker color indicates higher score.}
 		\label{figure:attn}
 	\end{figure}

		In order to intuitively interpret the strength of our proposed approach, we visualize the attention scores of self-attention layer to investigate whether the model has learned the crucial information that not exists in dependency tree of entities (i.e., \textit{partial response}) for the relation extraction. As shown in Fig.~\ref{figure:attn}, we observed that our ESC-GCN focus more attention on the center of the heat map, which means that higher scores are usually located in the middle of the sentence. Traditionally, words in the middle of the sentence are more likely to determine the relation of the entities. Besides, our model assigns the tokens (i.e., \textit{showed a partial response}) related to entities relatively higher scores, which helps to predict the gold relation.
	
	\input{table/prune.tex}

 	\section{Conclusion} \label{section:conclusion}
		In this paper, we propose a novel neural model for relation extraction that is based on graph convolutional networks over dependency trees. By incorporating the context of the words related to entities with inter-dependencies of input sentence, our model can capture the long-distance dependency relation between target entities more effectively, especially in long sentences. The experimental results demonstrate that our proposed model outperforms most baseline neural sequence-based models and dependency-based models. We further visualize the attention of our model to show how our relative position self-attention layer affects the model. In summary, our model effectively combines syntactic and semantic representations, which significantly improves the performance of relation extraction.

	\bibliographystyle{IEEEtran}
	\bibliography{IEEEabrv,ijcnn2021}
	
\clearpage
\input{appendix.tex}
\end{document}

%% file: table/crossdataset.tex
\begin{table}[tbp]
\centering
\begin{threeparttable}
\caption{Dataset statistics for Cross sentence $n$-ary task.}
\label{table:crossdataset}
	\begin{tabular}{lccc}
	\toprule 
	Dataset & Avg.Token & Avg.Sentence & Cross.Probability\\
	\midrule
	Ternary & 73.9 & 2.0 & 70.1\% \\
	Binary & 61.0 & 1.8 & 55.2\% \\
	\bottomrule
	\end{tabular}
``Avg.Token'' represents average number of words in sentence, ``Avg.Sentence'' represents average number of sentence in the instance, ``Cross.Probability'' represents cross percentage of dataset.
\end{threeparttable}
\end{table}

%% file: table/sentencedataset.tex
\begin{table}[tbp]
\centering
\begin{threeparttable}
\caption{Dataset statistics for sentence-level task.}
\label{table:sentencedataset}
	\begin{tabular}{lccc}
	\toprule 
	Dataset & Relation & Examples & Neg.examples\\
	\midrule
	TACRED & 42 & 106,264 & 79.5\% \\
	SemEval-2010 Task 8 & 19 & 10,717 & 17.4\% \\
	\bottomrule
	\end{tabular}
``Relation'' represents the number of relation type, ``Examples'' represents the number of sentence, ``Neg.examples'' represents the percentage of $no\_relation$.
\end{threeparttable}
\end{table}

%% file: table/cross.tex
\begin{table*}[thbp]
\centering
\begin{threeparttable}
\caption{Average test accuracies for binary-class $n$-ary relation extraction and multi-class $n$-ary relation extraction.}
\label{table:cross}
\setlength{\tabcolsep}{8pt}
\begin{tabular}{lcccccc} 
\toprule
    \multirow{3}{*}{\textbf{Model}} & \multicolumn{4}{c}{\textbf{Binary-class}} & \multicolumn{2}{c}{\textbf{Multi-class}} \\
\cmidrule(l{5pt}r{5pt}){2-5} \cmidrule(l{5pt}r{5pt}){6-7}
    & \multicolumn{2}{c}{Ternary} & \multicolumn{2}{c}{Binary}  & Ternary & Binary\\
& \texttt{Single} & \texttt{Cross} & \texttt{Single} & \texttt{Cross} & \texttt{Cross} & \texttt{Cross}  \\
\midrule
Feature-Based \cite{quirk2016distant} & 74.7 & 77.7 & 73.9 & 75.2 & - & - \\
SPTree \cite{miwa2016end} & - & - & 75.9 & 75.9 & - & - \\
\midrule Graph LSTM-EMBED \cite{peng2017cross} & 76.5 & 80.6 & 74.3 & 76.5 & - & - \\
Graph LSTM-FULL \cite{peng2017cross} & 77.9 & 80.7 & 75.6 & 76.7 & - & - \\
{\color{white}0000000000000000} + multi-task & - & 82.0 & - & 78.5 & - & - \\
Bidir DAG LSTM \cite{song2018n} & 75.6 & 77.3 & 76.9 & 76.4 & 51.7 & 50.7 \\
GS GLSTM \cite{song2018n} & 80.3 & 83.2 & 83.5 & 83.6 & 71.7 & 71.7 \\
\midrule GCN \cite{zhang2018graph} & 85.8 & 85.8 & 84.2 & 83.7 & 78.1 & 74.3 \\
\midrule ESC-GCN(ours) & \textbf{86.2} & \textbf{86.7} & \textbf{85.0} & \textbf{84.6} & \textbf{80.6} & \textbf{75.5} \\
\bottomrule
\end{tabular}
``Ternary'' and  ``Binary'' denote ternary drug-gene-mutation interactions and binary drug-mutation interactions, respectively. \texttt{Single} and \texttt{Cross} indicate that the entities of relations in single sentence or multiple sentences, respectively.
\end{threeparttable}
\end{table*}

%% file: table/tacred.tex
\begin{table}[tbp]
\centering
\caption{Micro-averaged precision (P), recall (R) and $F_{1}$ score on the TACRED dataset.}
\label{table:tacred}
	\begin{tabular}{lccc}
	\toprule
	\bf Model & P & R & $F_{1}$ \\
	\midrule
	CNN-PE \cite{nguyen2015relation} & 68.2 & 55.4 & 61.1 \\
	PA-LSTM \cite{zhang2017position} & 65.7 & 64.5 & 65.1 \\
	Self-Attn \cite{bilan2018position} & 64.6 & \textbf{68.6} & 66.5 \\
	SA-LSTM \cite{yu2019beyond} & 68.1 & 65.7 & 66.9 \\ 
	\midrule SDP-LSTM \cite{xu2015classifying} & 66.3 & 52.7 & 58.7 \\
	Tree-LSTM \cite{tai2015improved} & 66.0 & 59.2 & 62.4 \\
	GCN \cite{zhang2018graph} & 69.8 & 59.0 & 64.0 \\
	C-GCN \cite{zhang2018graph} & 69.9 & 63.3 & 66.4 \\
	S-GCN \cite{wu2019simplifying} & - &- & 67.0 \\
	\midrule ESC-GCN(ours) & \textbf{71.4} & 62.8 & \textbf{67.1} \\
	\bottomrule
	\end{tabular}
\end{table}

%% file: table/semeval.tex
\begin{table}[tbp]
	\centering
	\caption{Macro-averaged $F_{1}$ score on SemEval-2010 Task 8 dataset.}
	\label{table:semeval}
	\begin{tabular}{lr}
	\toprule 
	\textbf{Model} & Macro-$F_{1}$ \\
	\midrule
	CNN \cite{zeng2014relation} & 78.3 \\
	CR-CNN \cite{santos2015classifying} & 84.1 \\
	PA-LSTM \cite{zhang2017position} & 82.7 \\
	BiLSTM+Attn \cite{zhou2016attention} & 84.0 \\
	Entity-aware LSTM \cite{lee2019semantic} & \textbf{85.2} \\
	\midrule SDP-LSTM \cite{xu2015classifying} & 83.7 \\
	SPTree \cite{miwa2016end} & 84.4 \\
	C-GCN \cite{zhang2018graph} & 84.8 \\
	\midrule ESC-GCN(ours) & 85.1 \\
	\bottomrule
	\end{tabular}
\end{table}

%% file: table/ablation.tex
\begin{table}[tbp]
	\centering
    \caption{An ablation study for ESC-GCN model.}
    \label{table:ablation}
    \begin{tabular}{lc}
    	\toprule
    	\textbf{Model} & $F_{1}$ \\
    	\midrule
    	Best ESC-GCN & \textbf{67.1}\\
    	\quad-- Default Residual & 66.9 \\
    	\quad-- Layer Normalization & 66.7 \\
    	\quad-- Entity-aware Module & 66.6 \\
    	\quad-- Relative Position Self-Attention & 66.4 \\
    	\quad-- $h_{\text{s}}$, $h_{\text{o}}$, and Feedforward (FF) & 66.2\\
    	\quad-- BiLSTM Layer & 64.8 \\
    	\bottomrule
    \end{tabular}
\end{table}

%% file: table/prune.tex
\begin{table}[tbp]
	\centering
	\caption{Results of ESC-GCN with pruned trees.}
	\label{table:prune}
	\begin{tabular}{lcccc}
	\toprule
	\bf Model & & & & Dev-$F_{1}$ \\
	\midrule
	ESC-GCN (Full tree)         & & & & 66.0\\
	ESC-GCN ($K$=2)             & & & & 66.5\\
	ESC-GCN ($K$=1)             & & & & \textbf{67.1}\\
	ESC-GCN ($K$=0)             & & & & 66.3\\
	\bottomrule
	\end{tabular}
\end{table}

%% file: appendix.tex
\section*{Supplemental Material}
\label{sec:detail}

\subsection{Implementation Details}
We tune the hyper-parameters according to the validation sets. Following previous work \cite{zhang2018graph}, we employ the “entity mask” strategy where subject (object similarly) entity with special \textit{NER-$\langle$SUBJ$\rangle$} token, which can avoid overfitting and provide entity type information for relation extraction. We use the pre-trained 300-dimensional GloVe \cite{pennington2014glove} vectors as the initialization for word embeddings.\footnote{We use the 300-dimensional Glove word vectors, which is available at \url{http://nlp.stanford.edu/projects/glove/}} We randomly initialize all other embeddings (i.e., POS, NER, position embeddings) with 30-dimension vectors. We set LSTM hidden size to 200 in all neural models. We set self-attention hidden size to 130 and the number of heads to 3. We also use hidden size 200 for the GCN layers and the feedforward layers. We employ the ReLU function for all nonlinearities, and apply max-pooling operations in all pooling layers. We use the dependency parsing, POS and NER sequences in the original dataset, which was generated with Stanford CoreNLP \cite{manning2014stanford}. For regularization, we apply dropout with $p$ = 0.5 after the input layer and before the classifier layer. The model is trained using stochastic gradient descent optimizer for 100 epochs and decay rate of 0.9.

\subsection{Additional Analysis}

\begin{figure}
	\centering
	\includegraphics[width=0.43\textwidth]{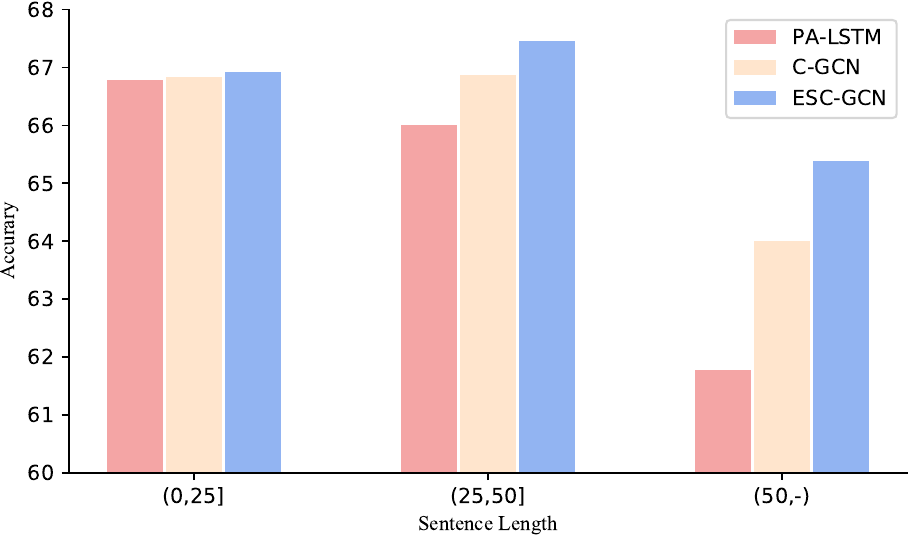}
	\caption{Test set performance with different sentence lengths for ESC-GCN, C-GCN and PA-LSTM. The PA-LSTM results are from \cite{zhang2017position}, and the C-GCN are from \cite{zhang2018graph}.}
	\label{figure:sentence}
\end{figure}

\textbf{Accuracy against sentence lengths}
To investigate the performance of our proposed model under different sentence lengths, we partition the test set of TACRED into three categories ((0, 25], (25, 50], (50,$\infty$]) with different sentence lengths. As shown in Fig.~\ref{figure:sentence}, we compare our ESC-GCN with C-GCN \cite{zhang2018graph} and PA-LSTM \cite{zhang2017position}. Obviously, our model outperforms C-GCN and PA-LSTM against various sentence lengths, showing the effectiveness of our model. Intuitively, longer instances are more challenging since it contains a lot of noise that is useless for relation extraction. However, we find that the performance gap is enlarged when the instance length increases, which demonstrates the superiority of our ESC-GCN in capturing long-distance dependencies.

\begin{figure}
	\centering
	\includegraphics[width=0.43\textwidth]{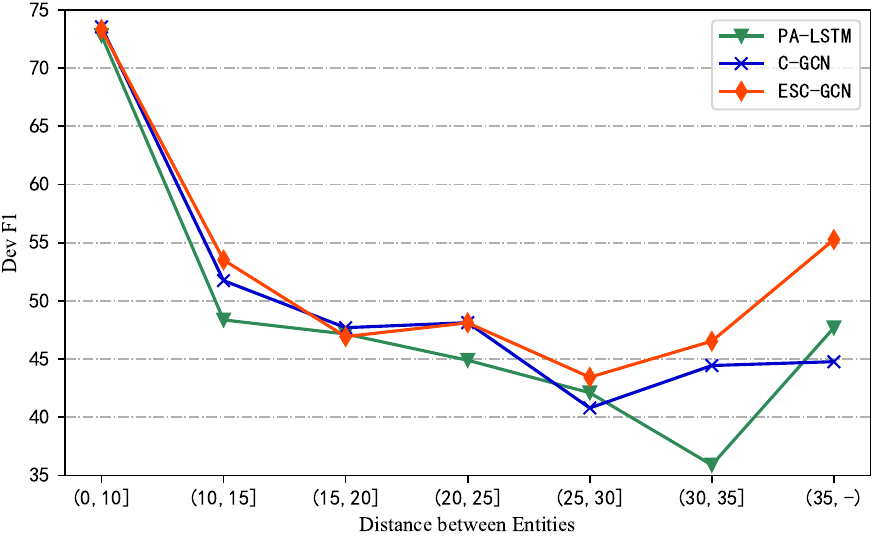}
	\caption{Dev set performance with different distance between entities for ESC-GCN, C-GCN and PA-LSTM. The PA-LSTM results are from \cite{zhang2017position}, and the C-GCN are from \cite{zhang2018graph}.}
	\label{figure:distance}
\end{figure}

\textbf{Accuracy against distance between entities}
We divide the dev set into seven types with different entity distance ((0, 10], (10, 15], (15, 20], (20, 25], (25, 30], (30, 35], (35,$\infty$]), and evaluate the performances of the PA-LSTM, C-GCN and ESC-GCN models. As shown in
Fig.~\ref{figure:distance}, our proposed model outperforms PA-LSTM for each set of instances, showing the effectiveness of dependency parsing for relation extraction. Compared with dependency-based model C-GCN, our ESC-GCN has consistent performance when the distance between entities is short. However, our ESC-GCN surpasses C-GCN much when the entity distance longer than 25. These results show that the superiority of ESC-GCN over C-GCN in utilizing semantic context related entities in terms of long distance between entities.

\begin{figure}
	\centering
	\includegraphics[width=0.43\textwidth]{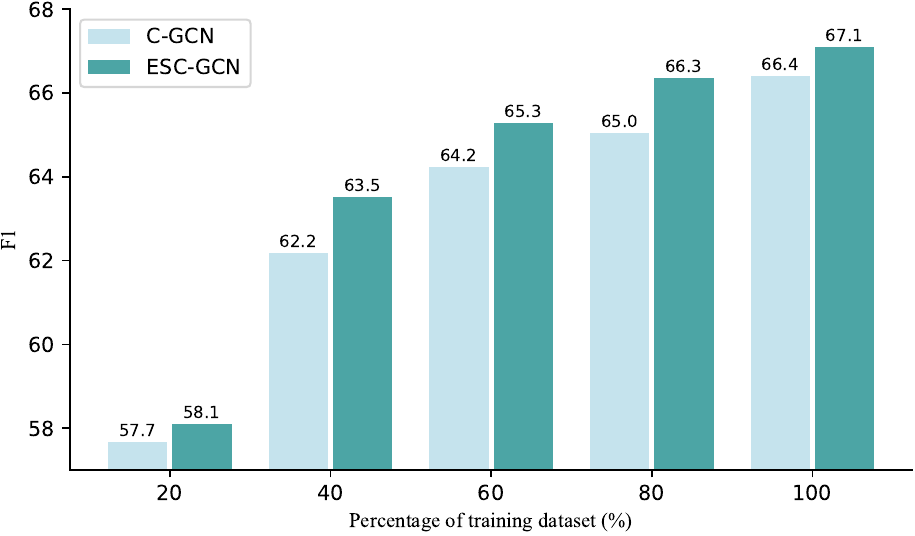}
	\caption{Comparison of ESC-GCN and C-GCN \cite{zhang2018graph} under different training data size.}
	\label{figure:datasize}
\end{figure}

\textbf{Accuracy against various training data sizes}
We set up five types of training data (20\%, 40\%, 60\%, 80\% and 100\%) and evaluate the performances of the C-GCN and ESC-GCN models. As shown in Fig.~\ref{figure:datasize}, our ESC-GCN consistently outperforms C-GCN under the same amount of training data. Generally, we find that when the size of the training data reaches to 40\%, our ESC-GCN has learned latent features related to relation extraction. Particularly, using 80\% of the training data, our model achieves $F_{1}$ value of 66.3, which closing to the best $F_{1}$ of C-GCN. These results show that our ESC-GCN can utilize training resources more effectively.

%% file: mypaper.bbl
\begin{thebibliography}{10}
\providecommand{\url}[1]{#1}
\csname url@samestyle\endcsname
\providecommand{\newblock}{\relax}
\providecommand{\bibinfo}[2]{#2}
\providecommand{\BIBentrySTDinterwordspacing}{\spaceskip=0pt\relax}
\providecommand{\BIBentryALTinterwordstretchfactor}{4}
\providecommand{\BIBentryALTinterwordspacing}{\spaceskip=\fontdimen2\font plus
\BIBentryALTinterwordstretchfactor\fontdimen3\font minus \fontdimen4\font\relax}
\providecommand{\BIBforeignlanguage}[2]{{%
\expandafter\ifx\csname l@#1\endcsname\relax
\typeout{** WARNING: IEEEtran.bst: No hyphenation pattern has been}%
\typeout{** loaded for the language `#1'. Using the pattern for}%
\typeout{** the default language instead.}%
\else
\language=\csname l@#1\endcsname
\fi
#2}}
\providecommand{\BIBdecl}{\relax}
\BIBdecl

\bibitem{yu2017improved}
M.~Yu, W.~Yin, K.~S. Hasan, C.~d. Santos, B.~Xiang, and B.~Zhou, ``Improved neural relation detection for knowledge base question answering,'' in \emph{Annual Meeting of the Association for Computational Linguistics}, 2017, pp. 571--581.

\bibitem{yin2022generic}
N.~Yin and Z.~Luo, ``Generic structure extraction with bi-level optimization for graph structure learning,'' \emph{Entropy}, vol.~24, no.~9, p. 1228, 2022.

\bibitem{sorokin2017context}
D.~Sorokin and I.~Gurevych, ``Context-aware representations for knowledge base relation extraction,'' in \emph{The Conference on Empirical Methods in Natural Language Processing}, 2017, pp. 1784--1789.

\bibitem{yin2022dynamic}
N.~Yin, F.~Feng, Z.~Luo, X.~Zhang, W.~Wang, X.~Luo, C.~Chen, and X.-S. Hua, ``Dynamic hypergraph convolutional network,'' in \emph{2022 IEEE 38th International Conference on Data Engineering (ICDE)}.\hskip 1em plus 0.5em minus 0.4em\relax IEEE, 2022, pp. 1621--1634.

\bibitem{qian2019graph}
Y.~Qian, E.~Santus, Z.~Jin, J.~Guo, and R.~Barzilay, ``Graphie: A graph-based framework for information extraction,'' in \emph{The Conference of the North American Chapter of the Association for Computational Linguistics}, 2019, pp. 751--761.

\bibitem{yin2022deal}
N.~Yin, L.~Shen, B.~Li, M.~Wang, X.~Luo, C.~Chen, Z.~Luo, and X.-S. Hua, ``Deal: An unsupervised domain adaptive framework for graph-level classification,'' in \emph{Proceedings of the 30th ACM International Conference on Multimedia}, 2022, pp. 3470--3479.

\bibitem{zeng2014relation}
D.~Zeng, K.~Liu, S.~Lai, G.~Zhou, and J.~Zhao, ``Relation classification via convolutional deep neural network,'' in \emph{International Conference on Computational Linguistics}, 2014, pp. 2335--2344.

\bibitem{zhang2015relation}
D.~Zhang and D.~Wang, ``Relation classification via recurrent neural network,'' \emph{arXiv:1508.01006}, 2015.

\bibitem{xu2015classifying}
Y.~Xu, L.~Mou, G.~Li, Y.~Chen, H.~Peng, and Z.~Jin, ``Classifying relations via long short term memory networks along shortest dependency paths,'' in \emph{The Conference on Empirical Methods in Natural Language Processing}, 2015, pp. 1785--1794.

\bibitem{miwa2016end}
M.~Miwa and M.~Bansal, ``End-to-end relation extraction using lstms on sequences and tree structures,'' in \emph{Annual Meeting of the Association for Computational Linguistics}, 2016, pp. 1105--1116.

\bibitem{zhang2017position}
Y.~Zhang, V.~Zhong, D.~Chen, G.~Angeli, and C.~D. Manning, ``Position-aware attention and supervised data improve slot filling,'' in \emph{The Conference on Empirical Methods in Natural Language Processing}, 2017, pp. 35--45.

\bibitem{lee2019semantic}
J.~Lee, S.~Seo, and Y.~S. Choi, ``Semantic relation classification via bidirectional lstm networks with entity-aware attention using latent entity typing,'' \emph{Symmetry}, vol.~11, no.~6, p. 785, 2019.

\bibitem{zhang2022would}
B.~Zhang, D.~Ding, L.~Jing, G.~Dai, and N.~Yin, ``How would stance detection techniques evolve after the launch of chatgpt?'' \emph{arXiv preprint arXiv:2212.14548}, 2022.

\bibitem{zhang2023investigating}
B.~Zhang, X.~Fu, D.~Ding, H.~Huang, G.~Dai, N.~Yin, Y.~Li, and L.~Jing, ``Investigating chain-of-thought with chatgpt for stance detection on social media,'' \emph{arXiv preprint arXiv:2304.03087}, 2023.

\bibitem{meng2024deep}
T.~Meng, Y.~Shou, W.~Ai, N.~Yin, and K.~Li, ``Deep imbalanced learning for multimodal emotion recognition in conversations,'' \emph{IEEE Transactions on Artificial Intelligence}, 2024.

\bibitem{wang2016relation}
L.~Wang, Z.~Cao, G.~De~Melo, and Z.~Liu, ``Relation classification via multi-level attention cnns,'' in \emph{Annual Meeting of the Association for Computational Linguistics}, 2016, pp. 1298--1307.

\bibitem{nguyen2015relation}
T.~H. Nguyen and R.~Grishman, ``Relation extraction: Perspective from convolutional neural networks,'' in \emph{The Workshop on Vector Space Modeling for Natural Language Processing}, 2015, pp. 39--48.

\bibitem{yin2024continuous}
N.~Yin, M.~Wan, L.~Shen, H.~L. Patel, B.~Li, B.~Gu, and H.~Xiong, ``Continuous spiking graph neural networks,'' \emph{arXiv preprint arXiv:2404.01897}, 2024.

\bibitem{santos2015classifying}
C.~N.~d. Santos, B.~Xiang, and B.~Zhou, ``Classifying relations by ranking with convolutional neural networks,'' in \emph{Annual Meeting of the Association for Computational Linguistics}, 2015, pp. 626--634.

\bibitem{shen2016attention}
Y.~Shen and X.-J. Huang, ``Attention-based convolutional neural network for semantic relation extraction,'' in \emph{International Conference on Computational Linguistics}, 2016, pp. 2526--2536.

\bibitem{zhang2015bidirectional}
S.~Zhang, D.~Zheng, X.~Hu, and M.~Yang, ``Bidirectional long short-term memory networks for relation classification,'' in \emph{Pacific Asia Conference on Language, Information and Computation}, 2015, pp. 73--78.

\bibitem{zhou2016attention}
P.~Zhou, W.~Shi, J.~Tian, Z.~Qi, B.~Li, H.~Hao, and B.~Xu, ``Attention-based bidirectional long short-term memory networks for relation classification,'' in \emph{Annual Meeting of the Association for Computational Linguistics}, 2016, pp. 207--212.

\bibitem{yin2024dynamic}
N.~Yin, M.~Wang, Z.~Chen, G.~De~Masi, H.~Xiong, and B.~Gu, ``Dynamic spiking graph neural networks,'' in \emph{Proceedings of the AAAI Conference on Artificial Intelligence}, vol.~38, no.~15, 2024, pp. 16\,495--16\,503.

\bibitem{zhang2018graph}
Y.~Zhang, P.~Qi, and C.~D. Manning, ``Graph convolution over pruned dependency trees improves relation extraction,'' in \emph{The Conference on Empirical Methods in Natural Language Processing}, 2018, pp. 2205--2215.

\bibitem{yinsport}
N.~Yin, L.~Shen, C.~Chen, X.-S. Hua, and X.~Luo, ``Sport: A subgraph perspective on graph classification with label noise,'' \emph{ACM Transactions on Knowledge Discovery from Data}.

\bibitem{ai2023gcn}
W.~Ai, Y.~Shou, T.~Meng, N.~Yin, and K.~Li, ``Der-gcn: Dialogue and event relation-aware graph convolutional neural network for multimodal dialogue emotion recognition,'' \emph{arXiv preprint arXiv:2312.10579}, 2023.

\bibitem{shou2023adversarial}
Y.~Shou, T.~Meng, W.~Ai, N.~Yin, and K.~Li, ``Adversarial representation with intra-modal and inter-modal graph contrastive learning for multimodal emotion recognition,'' \emph{arXiv preprint arXiv:2312.16778}, 2023.

\bibitem{tang2024merging}
A.~Tang, L.~Shen, Y.~Luo, N.~Yin, L.~Zhang, and D.~Tao, ``Merging multi-task models via weight-ensembling mixture of experts,'' in \emph{International Conference on Machine Learning}, 2024.

\bibitem{xu2016improved}
Y.~Xu, R.~Jia, L.~Mou, G.~Li, Y.~Chen, Y.~Lu, and Z.~Jin, ``Improved relation classification by deep recurrent neural networks with data augmentation,'' in \emph{International Conference on Computational Linguistics}, 2016.

\bibitem{ju2024survey}
W.~Ju, S.~Yi, Y.~Wang, Z.~Xiao, Z.~Mao, H.~Li, Y.~Gu, Y.~Qin, N.~Yin, S.~Wang \emph{et~al.}, ``A survey of graph neural networks in real world: Imbalance, noise, privacy and ood challenges,'' \emph{arXiv preprint arXiv:2403.04468}, 2024.

\bibitem{tai2015improved}
K.~S. Tai, R.~Socher, and C.~D. Manning, ``Improved semantic representations from tree-structured long short-term memory networks,'' in \emph{Annual Meeting of the Association for Computational Linguistics}, 2015, pp. 1556--1566.

\bibitem{peng2017cross}
N.~Peng, H.~Poon, C.~Quirk, K.~Toutanova, and W.-t. Yih, ``Cross-sentence n-ary relation extraction with graph lstms,'' in \emph{Transactions of the Association for Computational Linguistics}, vol.~5, 2017, pp. 101--115.

\bibitem{song2018n}
L.~Song, Y.~Zhang, Z.~Wang, and D.~Gildea, ``N-ary relation extraction using graph state lstm,'' in \emph{The Conference on Empirical Methods in Natural Language Processing}, 2018, pp. 2226--2235.

\bibitem{vashishth2018incorporating}
S.~Vashishth, M.~Bhandari, P.~Yadav, P.~Rai, C.~Bhattacharyya, and P.~Talukdar, ``Incorporating syntactic and semantic information in word embeddings using graph convolutional networks,'' \emph{arXiv:1809.04283}, 2018.

\bibitem{yin2023dream}
N.~Yin, M.~Wang, Z.~Chen, L.~Shen, H.~Xiong, B.~Gu, and X.~Luo, ``Dream: Dual structured exploration with mixup for open-set graph domain adaption,'' in \emph{The Twelfth International Conference on Learning Representations}, 2023.

\bibitem{sahu2020relation}
S.~K. Sahu, D.~Thomas, B.~Chiu, N.~Sengupta, and M.~Mahdy, ``Relation extraction with self-determined graph convolutional network,'' in \emph{ACM International Conference on Information \& Knowledge Management}, 2020, pp. 2205--2208.

\bibitem{vaswani2017attention}
A.~Vaswani, N.~Shazeer, N.~Parmar, J.~Uszkoreit, L.~Jones, A.~N. Gomez, {\L}.~Kaiser, and I.~Polosukhin, ``Attention is all you need,'' in \emph{Advances in Neural Information Processing Systems}, 2017, pp. 5998--6008.

\bibitem{yin2023omg}
N.~Yin, L.~Shen, M.~Wang, X.~Luo, Z.~Luo, and D.~Tao, ``Omg: towards effective graph classification against label noise,'' \emph{IEEE Transactions on Knowledge and Data Engineering}, 2023.

\bibitem{yin2023messages}
N.~Yin, L.~Shen, H.~Xiong, B.~Gu, C.~Chen, X.-S. Hua, S.~Liu, and X.~Luo, ``Messages are never propagated alone: Collaborative hypergraph neural network for time-series forecasting,'' \emph{IEEE Transactions on Pattern Analysis and Machine Intelligence}, 2023.

\bibitem{pang2023sa}
J.~Pang, Z.~Wang, J.~Tang, M.~Xiao, and N.~Yin, ``Sa-gda: Spectral augmentation for graph domain adaptation,'' in \emph{Proceedings of the 31st ACM International Conference on Multimedia}, 2023, pp. 309--318.

\bibitem{yin2023coco}
N.~Yin, L.~Shen, M.~Wang, L.~Lan, Z.~Ma, C.~Chen, X.-S. Hua, and X.~Luo, ``Coco: A coupled contrastive framework for unsupervised domain adaptive graph classification,'' in \emph{International Conference on Machine Learning}.\hskip 1em plus 0.5em minus 0.4em\relax PMLR, 2023, pp. 40\,040--40\,053.

\bibitem{tran2019relation}
V.-H. Tran, V.-T. Phi, H.~Shindo, and Y.~Matsumoto, ``Relation classification using segment-level attention-based cnn and dependency-based rnn,'' in \emph{The Conference of the North American Chapter of the Association for Computational Linguistics}, 2019, pp. 2793--2798.

\bibitem{verga2018simultaneously}
P.~Verga, E.~Strubell, and A.~McCallum, ``Simultaneously self-attending to all mentions for full-abstract biological relation extraction,'' in \emph{The Conference of the North American Chapter of the Association for Computational Linguistics}, 2018, pp. 872--884.

\bibitem{bilan2018position}
I.~Bilan and B.~Roth, ``Position-aware self-attention with relative positional encodings for slot filling,'' \emph{arXiv:1807.03052}, 2018.

\bibitem{yu2019beyond}
B.~Yu, Z.~Zhang, T.~Liu, B.~Wang, S.~Li, and Q.~Li, ``Beyond word attention: Using segment attention in neural relation extraction,'' in \emph{International Joint Conference on Artificial Intelligence}, 2019, pp. 5401--5407.

\bibitem{kipf2016semi}
T.~N. Kipf and M.~Welling, ``Semi-supervised classification with graph convolutional networks,'' in \emph{International Conference on Learning Representations}, 2017.

\bibitem{marcheggiani2018exploiting}
D.~Marcheggiani, J.~Bastings, and I.~Titov, ``Exploiting semantics in neural machine translation with graph convolutional networks,'' in \emph{The Conference of the North American Chapter of the Association for Computational Linguistics}, 2018, pp. 486--492.

\bibitem{shaw2018self}
P.~Shaw, J.~Uszkoreit, and A.~Vaswani, ``Self-attention with relative position representations,'' in \emph{The Conference of the North American Chapter of the Association for Computational Linguistics}, 2018, pp. 464--468.

\bibitem{liu2015dependency}
Y.~Liu, F.~Wei, S.~Li, H.~Ji, M.~Zhou, and H.~Wang, ``A dependency-based neural network for relation classification,'' in \emph{Annual Meeting of the Association for Computational Linguistics}, 2015, pp. 285--290.

\bibitem{quirk2016distant}
C.~Quirk and H.~Poon, ``Distant supervision for relation extraction beyond the sentence boundary,'' in \emph{The Conference of the European Chapter of the Association for Computational Linguistics}, 2016, pp. 1171--1182.

\bibitem{hendrickx2019semeval}
I.~Hendrickx, S.~N. Kim, Z.~Kozareva, P.~Nakov, D.~O. S{\'e}aghdha, S.~Pad{\'o}, M.~Pennacchiotti, L.~Romano, and S.~Szpakowicz, ``Semeval-2010 task 8: Multi-way classification of semantic relations between pairs of nominals,'' in \emph{The International Workshop on Semantic Evaluation}, 2019, pp. 33--38.

\bibitem{wu2019simplifying}
F.~Wu, T.~Zhang, A.~H.~d. Souza~Jr, C.~Fifty, T.~Yu, and K.~Q. Weinberger, ``Simplifying graph convolutional networks,'' in \emph{International Conference on Machine Learning}, 2019.

\bibitem{pennington2014glove}
J.~Pennington, R.~Socher, and C.~D. Manning, ``Glove: Global vectors for word representation,'' in \emph{The Conference on Empirical Methods in Natural Language Processing}, 2014, pp. 1532--1543.

\bibitem{manning2014stanford}
C.~D. Manning, M.~Surdeanu, J.~Bauer, J.~R. Finkel, S.~Bethard, and D.~McClosky, ``The stanford corenlp natural language processing toolkit,'' in \emph{Annual Meeting of the Association for Computational Linguistics}, 2014, pp. 55--60.

\end{thebibliography}
